\def\extended{true}

\documentclass[letterpaper, 10 pt, conference]{ieeeconf}

\IEEEoverridecommandlockouts                              

\usepackage[style=ieee]{biblatex}

\usepackage{adjustbox}
\usepackage{graphicx} \usepackage{amsmath} \usepackage{mathtools}
\usepackage{amssymb}
\usepackage{subfigure}
\usepackage[ruled,vlined,linesnumbered]{algorithm2e}

\usepackage{xr}
\usepackage[hidelinks]{hyperref}

\AtBeginDocument{
  \label{CorrectFirstPageLabel}
  
}

\usepackage[noabbrev]{cleveref}
\DeclareRobustCommand{\abbrevcrefs}{\Crefname{appendix}{App.}{Apps.}\Crefname{section}{Sec.}{Secs.}\Crefname{equation}{Eq.}{Eqs.}\Crefname{figure}{Fig.}{Figs.}\Crefname{algorithm}{Alg.}{Algs.}\Crefname{tabular}{Tab.}{Tabs.}\Crefname{lemma}{Lem.}{Lems.}\Crefname{corollary}{Cor.}{Cors.}\Crefname{theorem}{Thm.}{Thms.}\Crefname{proposition}{Prop.}{Props.}\Crefname{line}{L.}{Ls.}\crefname{appendix}{app.}{apps.}\crefname{section}{sec.}{secs.}\crefname{equation}{eq.}{eqs.}\crefname{figure}{fig.}{figs.}\crefname{algorithm}{alg.}{algs.}\crefname{tabular}{tab.}{tabs.}\crefname{lemma}{lem.}{lems.}\crefname{corollary}{cor.}{cors.}\crefname{theorem}{thm.}{thms.}\crefname{proposition}{prop.}{props.}\crefname{line}{l.}{ls.}}
\DeclareRobustCommand{\cshref}[1]{{\abbrevcrefs\cref{#1}}}
\DeclareRobustCommand{\Cshref}[1]{{\abbrevcrefs\Cref{#1}}}

\crefalias{AlgoLine}{line}

\usepackage{float}
\usepackage{booktabs}
\usepackage{array}
\usepackage{siunitx}

\usepackage{etoolbox}
\renewcommand{\bfseries}{\fontseries{b}\selectfont}
\newrobustcmd{\B}{\bfseries}

\usepackage[table]{xcolor}

\usepackage{comment}

\usepackage{calc}
\newcommand{\varendash}[1][5pt]{\makebox[#1]{\leaders\hbox{--}\hfill\kern0pt}}
\newcommand{\lineOf}[1]{\varendash[\widthof{#1}]}

\let\labelindent\relax
\usepackage[inline]{enumitem} 

\usepackage[normalem]{ulem}
\usepackage{scalefnt}

\newtheorem{definition}{Definition}

\newcommand{\eqdef}     {\stackrel{{\textrm{\rm\tiny def}}}{=}}

\newcommand{\persComment}[3]{
  \ifmmode
  \text{\textcolor{#3}{[#2] \em #1}}
  \else
  \textcolor{#3}{[#2] \em #1}
  \fi
}

\def\ie{{\em i.e.}\xspace}
\def\eg{{\em e.g.}\xspace}
\def\cf{{\em cf.}\xspace}

\def\reals{{\mathbb R}}
\def\cal{\mathcal}
\def\cS{{\cal S}}
\def\cA{{\cal A}}

\def\cI{{\cal I}}

\def\cE{{\cal E}}
\def\fsc{\mathit{fsc}}
\newcommand{\angles}[1]{\langle #1 \rangle}

\title{\LARGE \bf
Robust Robot Planning for Human-Robot Collaboration
}

\author{Yang You$^{1}$, Vincent Thomas$^{1}$, Francis Colas$^{1}$, Rachid Alami$^{2}$, and Olivier Buffet$^{1}$\thanks{This work was supported by the French National Research
    Agency (ANR) through the “Flying Coworker” Project under Grant
    18-CE33-0001.}
\thanks{$^{1}$ Université de Lorraine, INRIA, CNRS, LORIA,
    F-54000 Nancy, France {\tt\small firstname.lastname@loria.fr}}
\thanks{$^{2}$ LAAS-CNRS, Université de Toulouse, CNRS,
    F-31000 Toulouse, France {\tt\small firstname.lastname@laas.fr}}
}

\usepackage{xcolor}
\newcommand{\Vincent}[1]{\textcolor{blue}{\textbf{[VT]}#1\textbf{[/VT]}}}
\newcommand{\vincent}[1]{\Vincent{#1}}

\newcommand{\Olivier}[1]{\textcolor{teal}{\textbf{[ob]} #1 \textbf{[/ob]}}}

\newcommand{\olivier}[1]{\Olivier{#1}}

\newcommand{\Yang}[1]{\textcolor{brown}{\textbf{[YY]}#1\textbf{[/YY]}}}
\newcommand{\yang}[1]{\Yang{#1}}

\begin{document}

\maketitle
\thispagestyle{empty}
\pagestyle{empty}

\begin{abstract}
In human-robot collaboration, the objectives of the human are often unknown to the robot.
Moreover, even assuming a known objective, the human behavior is also uncertain.
In order to plan a robust robot behavior, a key preliminary question is then:
How to derive realistic human behaviors given a known
  objective?
A major issue is that such a human behavior should itself account for the robot behavior, otherwise collaboration cannot happen.
In this paper, we rely on Markov decision models, representing
  the uncertainty over the human objective as a probability
  distribution over a finite set of
  objective functions
  (inducing a distribution over human behaviors).
Based on this, we propose two contributions:
\begin{enumerate*}
  \item an approach to automatically generate an uncertain human behavior (a policy) for each given objective function while accounting for possible robot behaviors;
and
  \item a robot planning algorithm that is robust to the above-mentioned uncertainties and relies on solving a partially observable Markov decision process (POMDP) obtained by reasoning on a distribution over human behaviors.
  \end{enumerate*}
A co-working scenario allows conducting experiments and presenting
  qualitative and quantitative results to evaluate our approach.
\end{abstract}

\section{INTRODUCTION}

Building smart robots to assist a human partner is a topical subject with applications in
manufacturing \cite{manufacturing_hrc, icra_2015, WANG20175, rsj2013}, healthcare \cite{acm2013}, etc.
In those applications, the robot often adapts to a single fixed human objective.
But to make it robust, we need to consider how the robot could adapt if the human's objective and his induced behavior are uncertain.

To circumvent the uncertainty over human objectives, \citeauthor*{cirl_NIPS2016} \cite{cirl_NIPS2016} proposed 
the CIRL framework (cooperative inverse reinforcement learning), where both the human and the robot have to maximize the human's reward function, which is hidden to the robot.
This reward function can be seen as encoding the task of the human but also, via intermediary rewards, his preferences.
In the CIRL framework, solving the planning problem is done through seeking a pair of behaviors, or policies, one for the robot and one for the human, which optimize the reward in the long term.
However, CIRL relies on two strong assumptions we want to relax:
\begin{enumerate*}
\item The human should follow the policy computed for him, while it may often be too complex to communicate or learn.
\item All state variables are visible, except for the human's objective (reward function parameter), which is hidden to the robot.
\end{enumerate*}

In this work, we want to compute a robot policy robust to incomplete information, and in particular on the human objectives and preferences, but also to unplanned behavior.
To consider an independent human in this robot planning problem, we need a model of his behavior that accounts for the robot's possibility to collaborate.
This induces a chicken-and-egg issue as the robot behavior is unknown at this stage.

Our first contribution consists in modeling a \emph{collaborative} human behavior from a given reward function.
This is done by \begin{enumerate*}
\item {\em temporarily} assuming that the human can control the robot with direct access to its observations, to compute the optimal value function of the resulting (multiagent) partially-observable Markov decision process (POMDP);
\item then, relaxing the shared observability assumption, extracting a model for the human alone from that value function.
\end{enumerate*}

Our second contribution is to derive and solve a planning problem to build a robust policy for the robot.
This is achieved by constructing a new POMDP including, as part of the robot environment, a model of the human as a global stochastic finite-state controller (FSC), which samples the FSCs extracted for each given reward function as above.

\Cref{sec:RelatedWork} discusses related works in human-robot collaboration.
\Cshref{sec:Background} formally defines POMDPs, Decentralized POMDPs, and FSCs.
\Cshref{sec:Generate_human_policies} explains how to automatically generate a stochastic human FSC for a given reward function.
\Cshref{sec:robust_best_response} describes how to build a robust robot policy to adapt to several possible human policies.
Finally, \Cshref{sec:experiments} presents empirical results obtained with both synthetic and real humans on a high-level task in a simulated environment.

\ifdefined\extended
Note: This extended version provides additional details and results in its appendix.
\else
Note: Additional details and results are provided in the appendix of an extended version of this paper \cite{extended}.
\fi

\section{Related Work}
\label{sec:RelatedWork}

One can distinguish between different approaches to human-robot collaboration depending on the (robot's) ``human mental model'', which, according to \citeauthor*{HRI_human_model_survey} \cite{HRI_human_model_survey}, can be in one of the three following categories:
\begin{description*}
\item{\em first-order mental model (1oMM)}: the robot considers that the human behaves independently and does not account for the robot's possible actions;
\item{\em second-order mental model (2oMM)}: the robot considers that the human accounts for the robot, which induces some form of recursive modeling up to a certain depth;
\item{\em shared-mental model (SMM)}: an SMM assumes that all agents in the team have common expectations, thus reason in the same manner, which ensures an optimal coordination.
\end{description*}
Of course this categorization symmetrically applies to how the human models the robot.
Let us mention that mental models can also been used in other AI planning settings, \eg, related to explainability \cite{SreChaMuiKam-aaai20}.

We focus here on problems with stochastic dynamics and
partial observability, which leads us to considering Markov decision
models. In this setting, a 1oMM typically corresponds to a POMDP where the
objective is to find the policy of one agent of interest, while the (a
priori known) policy of the other agent is part of the system
dynamics.
For instance, 
a ``robot POMDP'' assuming a known human behavior is solved in
\cite{HRI_2020, Nikolaidis_2017, Nikolaidis_2018, Nikolaidis_2020}.
SMMs can be formalized as {\em Decentralized POMDPs}
(Dec-POMDPs), typically used to optimize the joint
policy of a team of agents (with a common reward function).
CIRL can be seen as a special case where the human has full
observability and the robot's only hidden variable corresponds to
the actual human objective (his reward function), which allows for
dedicated solution techniques close to solving a POMDP.
For their part, 2oMMs can
be formalized as {\em Interactive POMDPs} (I-POMDPs) \cite{IPOMDP},
where agents model each others in a
  nested manner.

1oMMs will fail in many tasks requiring an explicit collaborative
behavior from the human, and
the SMMs' main assumption is typically too strong when collaborating
with a human.
We will thus equip the robot with a 2oMM of the human.
Our approach to robust robot planning could be formalized as an
I-POMDP, what we avoid mainly to simplify notations.
However, 2oMMs always raise a chicken-and-egg problem as deriving the required human behavior implies reasoning about the robot behavior we are trying to derive in the first place.
This is true for instance in case of
\begin{enumerate*}
\item a hand-made human policy \cite{HRI_2020, Nikolaidis_2017}, where the human designer has to reason on the robot behavior;
\item a learned human policy \cite{Zheng_CDC2018, Unhelkar_Shah_2019} or human reward (through IRL) \cite{Russell98, Ng00_IRL}, which requires a collaborative robot in the first place to demonstrate a collaborative behavior;
or
\item a planned human policy: the model needs to include the robot behavior.
\end{enumerate*}

In this work, we generate plausible human behaviors---which will serve as the robot's mental model of the human---through planning, and address the chicken-and-egg problem through answering the question:
"{\em What if the human could also control the robot?}", thus temporarily assuming that human and robot share their observations, which amounts to the human adopting a particular SMM.

\section{Background}
\label{sec:Background}

\subsection{Dec-POMDPs}

We use Dec-POMDPs only to formalize the collaboration problem for convenience,
but do not solve a Dec-POMDP to get a joint policy for the human and the robot.

\begin{definition}
  A {\em Dec-POMDP} with $|\cI|$ agents is defined by a tuple $M \equiv \langle \cI, \cS, \cA, \Omega, T, O, R, b_0, \gamma \rangle$, where:
$\cI = \{1, \dots, |\cI|\}$ is a set of {\em agents};
$\cS$ is a set of {\em states};
$\cA = \bigtimes_i \cA^i$ is a set of joint actions, with $\cA^i$ the set of agent $i$'s {\em actions}; $\Omega = \bigtimes_i \Omega^i$ is a set of joint
  observations, with $\Omega^i$ the set of  $i$'s {\em observations}; $T: \cS \times \cA \times \cS \to \reals$ is the {\em transition
    function}, with $T(s,a,s')$ the probability of transiting from $s$ to $s'$ if $a$ is
  performed;
$O: \cA \times \cS \times \Omega \to \reals$ is the
  {\em observation function}, with $O(a,s',o)$ the probability of observing $o$ if $a$ is performed and
  the next state is $s'$;
$R: \cS \times \cA \to \mathbb{R}$ is the {\em reward function},
  with $R(s,a)$ the immediate reward for performing $a$ in $s$;
$b_0$ is the {\em initial probability distribution} over states;
and $\gamma \in [0,1)$ is the {\em discount factor} applied to future rewards.
\end{definition}

An agent's $i$ action {\em policy} $\pi^i$ maps its possible
action-observation histories to distributions over actions.
The objective is then to find a joint policy
$\pi \equiv \langle \pi^1, \dots, \pi^{|\cI|} \rangle$ that maximizes the expected discounted return from $b_0$:
\begin{align*}
  V^{\pi}(b_0)
  & \eqdef \mathbb{E}\left[ \sum_{t=0}^{\infty} \gamma^{t} r(S_t, A_t) \mid S_0 \sim b_0, \pi \right].
\end{align*}

In a POMDP, \ie, when $\cI= \{ 1 \}$, many solvers rely on estimating
the optimal value function $V^*(b)$, or the action-value 
$Q^*(b,a)\eqdef R(b,a) + \gamma \sum_{o} Pr(o|b,a) \cdot V^*(b^o_a)$,
  where $b^o_a$ is the next belief when performing $a$ and observing
  $o$.

\subsection{Finite State Controllers}
\label{sec:FSC_def}

In this work, human policies are presented in the form of {\em finite state controllers} (FSC) (also called {\em policy graphs} \cite{MeuKimKaeCas-uai99}), \ie, automata 
which contain in each internal state a probability distribution from which to sample an action, and
whose transitions from one internal state to the next depend on the action performed and observation received.

\begin{definition}
  For some POMDP sets $\cA$ and $\Omega$, a stochastic {\em FSC} 
  is defined by a tuple
  $\fsc \equiv \langle N, \beta, \eta, \psi \rangle$, where:
  \begin{itemize}
  \item $N$ is a finite set of (internal) nodes;\item $\beta$ is a probability distribution from which to sample the initial node;
  \item $\eta: N \times \cA \times \Omega \times N \to \reals$, the transition
    function, gives the probability $\eta(n, \langle a, o \rangle,n')$ of transiting from
    node $n$ to $n'$ if $a$ is performed and $o$ is observed; 
a deterministic transition can be noted $n'=\eta(n, \langle a, o \rangle)$;
\item $\psi: N \times \cA \to \reals$, the action selection
    function, gives the probability $\psi(n,a)$ of choosing 
    $a \in \cA$ when in $n$. \end{itemize}
\end{definition}

\section{Generating Human Policies with Objectives}
\label{sec:Generate_human_policies}

We now describe how, from a known reward function (a human objective), to generate one of the human FSCs which will feed the robot planner in \Cshref{sec:robust_best_response}.
To model uncertainty, the process can be tuned to create more or less erratic human behaviors.\footnote{Uncertainty about human objectives is handled in \Cshref{sec:robust_best_response}.}

The reward function at hand induces a Dec-POMDP $D$ describing a collaboration problem.
As stated before, to account for possible interactions, an issue is that the human policy we are seeking depends on the robot policy we don't have in the first place.
To overcome this chicken-and-egg problem, we temporarily assume that the human can control the robot's actions and has access to its observations.
The Dec-POMDP is thus first relaxed as an MPOMDP $M$ (Multi-agent POMDP) \cite{Pynadath-jair02}, \ie, a single-agent problem. 
$M$ can be fed to a POMDP solver to compute an optimal action-value $Q^*_M(b,a)$ for any (belief,joint-action) pair $(b,a)$.

To derive a human policy usable in Dec-POMDP $D$, \ie, mapping human action-observation histories (alone) to human actions, let us now assume
\begin{enumerate*}
\item that the human does not control the robot anymore, but
\item that the robot still has access to the same belief $b$ as inferred by the human.
\end{enumerate*}
Given a belief $b$, we can model the uncertainty over human (and robot) behaviors using a softmax function over action-values to obtain a distribution over multiple optimal or near-optimal joint actions: $f(a|T,b) = {e^{\frac{Q^*_M(b,a)}{T}}} / {\sum_{a'} e^{\frac{Q^*_M(b,a')}{T}}}$,
where $T>0$ is a temperature parameter that makes the human
more {\em rational} if $T$ is low, only optimal actions being
selected, and
more {\em erratic} if $T$ is high, with a distribution close to
uniform.
Then, because both are now independent, human action $a_H$ is sampled with probability $\sigma_H^T(a_H|b) \eqdef \sum_{a_R} f(a_H,a_R|T,b)$, and robot action $a_R$ is sampled w.p. $\sigma_R^T(a_R|b) \eqdef \sum_{a_H} f(a_H,a_R|T,b)$.
Given these two action-sampling rules $\sigma_H^T$ and $\sigma_{R}^T$, the human can update his belief in Dec-POMDP $D$ given his last pair $(a_H,o_H)$ by marginalizing over the robot's private actions and observations.

Using these processes to pick a human action and to update his belief, we designed an algorithm that recursively extracts a human policy (represented as an FSC) building on \citeauthor*{YouThoColBuf-ictai2021}'s Algorithm~2 for standard POMDPs \cite{YouThoColBuf-ictai2021} (see also \citeauthor*{GrzPouYanHoe-ToCyb14}'s work \cite{GrzPouYanHoe-ToCyb14}).
In our setting, due to the stochastic decisions, the FSC transitions depend not only on the last observation, but also on the last action.
Note that our algorithm allows trading off the FSC quality with its complexity through
\begin{enumerate*}
\item bounding the FSC's number of nodes by $N_{\max}$, and
\item merging nodes when their reference beliefs (the beliefs used when creating the nodes) are within $\epsilon$ of each other.
\end{enumerate*}
Also, inspired by LAO* \cite{HANSEN200135}, we obtain a better FSC (as confirmed through experiments) by expanding, at each iteration, the (open) node which seems to contribute most to $b_0$'s value.
To that end, we select the open node $n$ that has the highest estimated value $V^*(n.b)$, weighted by the probability to reach that node.
Finally, self-loops are added in each node for human observations that are impossible under the current belief,
so that the resulting human FSC can be used whatever the robot's actual behavior.

This approach (further detailed in \Cshref{ext-app|FscExtraction}) extracts a bounded-size stochastic FSC encoding a variety of human behaviors.
In our experiments, all $Q^*(b, a)$ and $V^*(b)$ values are estimated using
\begin{enumerate*}
\item SARSOP for offline pre-computations \cite{sarsop}, and
\item POMCP to obtain good estimates online quickly, even in beliefs not visited by an optimal policy \cite{NIPS2010_edfbe1af}.
\end{enumerate*}

{\em Deterministic FSCs} will also be generated to simulate various humans in our experiments.
This is achieved by replacing each node's distribution over human actions by a single action sampled from that distribution.
See \Cshref{ext-app:samplingDeterministicFSC}.

\section{Robust Robot Task Planning}
\label{sec:robust_best_response}

We now want to derive a robot policy that is robust to different possible (hidden) human objectives, each attached to a different behavior.
This problem is formalized as a Dec-POMDP $D$, except that \begin{enumerate*}
\item the exact reward function (among $\rho$ candidates) is only known by the human; 
\item each reward function $r_i$ is attached to a human  FSC $\fsc_i \equiv \langle N_i, \beta_i, \eta_i, \psi_i \rangle$ (\cf \Cshref{sec:Generate_human_policies}); and 
\item the robot is given a probability distribution over the possible reward-FSC pairs: $P(\{(r_1, \fsc_1) ,\dots, (r_\rho, \fsc_\rho) \})$.
\end{enumerate*}
As detailed below, this robust robot behavior is obtained by first turning this distribution over FSCs into a single FSC, then using this FSC to derive a POMDP, and finally solving this POMDP.

To turn this probability distribution over human FSCs into a single FSC, we take their ``union'', the new distribution over initial nodes amounting to
\begin{enumerate*}
\item sampling one FSC $\fsc_i$ from the distribution $P( \{ \fsc_1, \dots, \fsc_\rho \} )$, and then
\item sampling a node from $\beta_i$.
\end{enumerate*}
More formally, the {\em union FSC} is defined as:
\begin{align*}
  N & \eqdef \bigcup_{i=1}^\rho N_i, \\
\beta(n) & \eqdef \beta_{i(n)}(n) \cdot P(\fsc_{i(n)}),
\intertext{where $i(n)$ is the id of $n$'s parent FSC: $i(n) \eqdef i \text{ s.t. } n\in N_i$,}
  \eta(n,\langle a,o \rangle,n') & \eqdef \begin{cases}
    \eta_{i(n)}(n,\langle a,o \rangle,n') & \text{if } n'\in N_{i(n)}, \\
0 & \text{otherwise,}
  \end{cases}
  \\
  \text{and }
  \psi(n,a) & \eqdef \psi_{i(n)}(n,a).
\end{align*}

Given this FSC, we can now formalize the robot's decision problem as a POMDP  where each {\em extended state} $e^t \in \cE$ contains a current world state $s^t$, a robot current observation $o^t_R$ and the current node $n^{t}_H$ inside the human union FSC:
$ e^t  \equiv \langle s^t, n^{t}_H, o^t_R \rangle. $

Based on the Dec-POMDP $D$ and on the associated union FSC, the dynamics and the reward function of the robot's decision problem can be written as follows:
\begin{align*}
   & \hspace{0em}  T_e(e^{t+1}, e^t, a^t_R)  = Pr(e^{t+1}| e^t, a^t_R) \\
& = \sum_{a^t_H} \sum_{o^{t+1}_H} T(s^t, \langle a^t_H, a^t_{R} \rangle, s^{t+1}) \cdot \eta(n^t_H,  \langle a^t_{H}, o^{t+1}_H \rangle, n^{t+1}_H)\cdot \\
& \qquad O(s^{t+1},  \langle a^t_H, a^t_{R} \rangle, \langle o^{t+1}_{H},o^{t+1}_R \rangle) \cdot \psi(n^t_H, a^t_H),
\\ 
 & \hspace{0em} O_e(e^{t+1},a^t_R,o^{t+1}_R)  = Pr(o^{t+1}_R|e^{t+1},a^t_R) = \mathbf{1}_{o^{t+1}_R = \tilde  o^{t+1}_R} \intertext{(where $\tilde{o}^{t+1}_R$ is the observation in $e^{t+1}$), and}
& r_e(e^t, a^t_R)
 = \sum_{a^t_H} r_{i(n_H^t)} (s^t, \langle a^t_H,a^t_R \rangle ) \cdot \psi(n^t_H, a^t_H).
\end{align*}
Solving this robot POMDP gives a robust robot policy which is a best response to the provided probability distribution over human policies.

\section{Experiments}
\label{sec:experiments}

\subsection{Experimental Setting}

Our experiments have been conducted on a laptop with an 2.3 GHz i9 cpu. The source code will be made available under MIT license.

\begin{figure}\centering
     \includegraphics[width=0.66\linewidth]{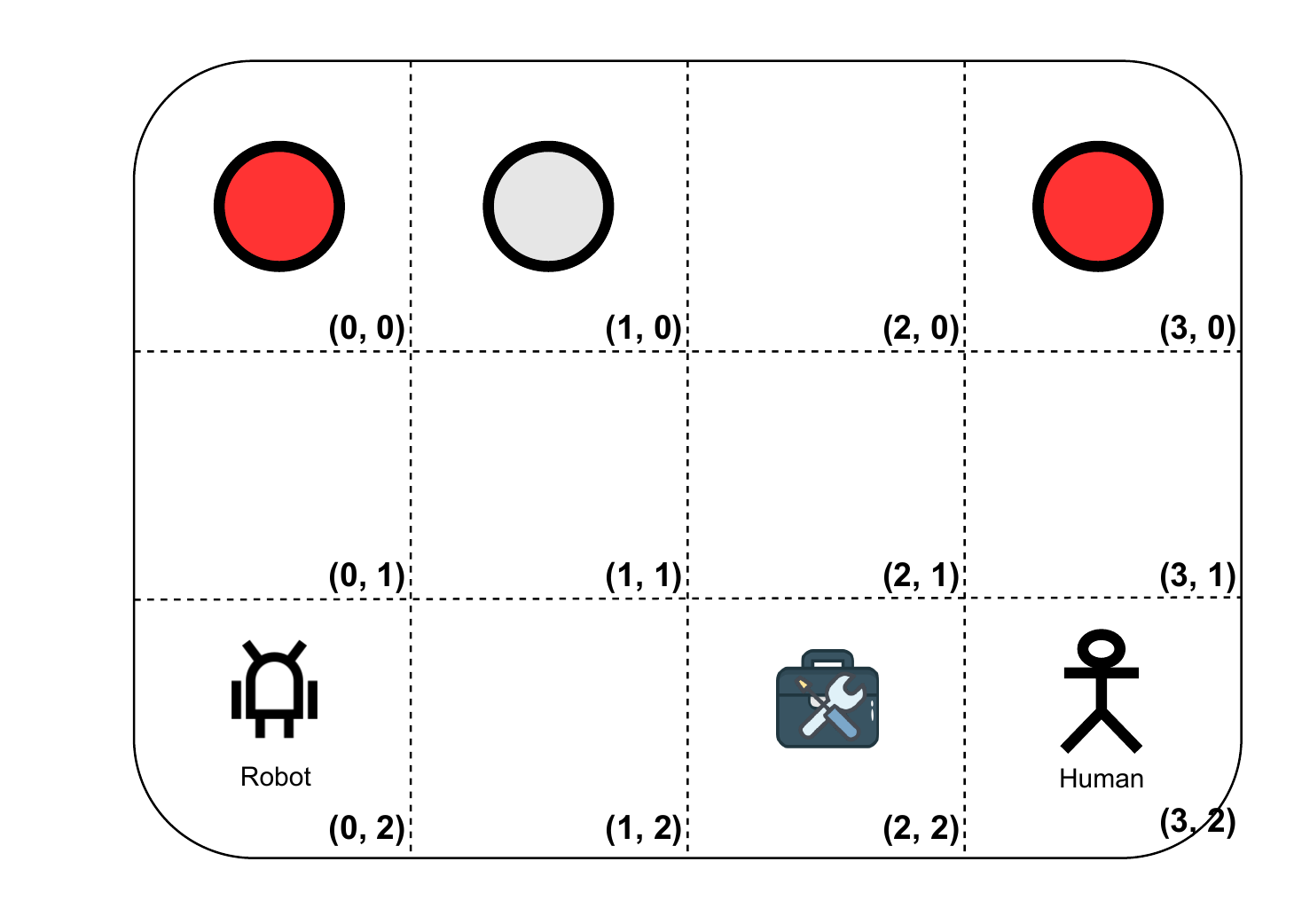}
     \caption{A collaboration task:
A robot and a human evolve in a $4\times 3$ grid world. 
The top-left cell $(0,0)$ and the top-right cell $(3,0)$ both contain a broken device.
A device to be maintained is also located in cell $(1,0)$.
A toolbox is located in cell $(2,2)$ where the human can pick components.
}
     \label{Collaboration_Task}
\end{figure}

To test our approach, we have designed a scenario presented in \Cref{Collaboration_Task},
where a robot and a human have to repair and maintain several devices located in a grid world.
One device needs to be maintained by the robot, which should be on the device's cell.
Repairing one of the two broken devices requires both the human and the robot to perform repair actions simultaneously at the device location, the human having previously picked a component in a toolbox.
Note also that both the human and the robot have limited observation of the environment.

This task is specified as the following Dec-POMDP:
\begin{itemize}
\item \uline{States ($S$):} The state $s \in S$ of the problem is made up of:
  \begin{enumerate*}
  \item the human location,
  \item the robot location,
  \item the status of the devices, and
  \item whether the human has a component or not.
  \end{enumerate*}
The human and robot locations are represented by integer coordinates $(x,y)$. They can be on the same cell. 
The state of each device is either "good", "broken", or "needs maintenance".
    
    \item \uline{Human observations ($\Omega_{H}$):}
      The human observes
      \begin{enumerate*}
        \item his location;
        \item whether the robot is on his cell or not; 
        \item the status of the device in his cell (if any).
        \end{enumerate*}
    
    \item \uline{Robot observations ($\Omega_{R}$):}
      The robot observes
      \begin{enumerate*}
      \item its location;
      \item the human's location; 
      \item the status of the device in its cell (if any).
      \end{enumerate*}
   
    \item \uline{Actions and Dynamics:}
    Both the human and the robot have the following actions:
{\em Up, Down, Left, Right} are the agent's 4 move actions;
{\em Wait}: the agent stays in his current position;
{\em Repair}: repairs a broken device if:
\begin{enumerate*}
    \item the human holds a new component; 
    \item the human and the robot are in the same cell as the broken device;
    \item the human and the robot both perform the repair action.
    \end{enumerate*}
Upon success, 
    the broken device turns to {\em good} and the human's component is consumed.

The human can {\em Pick a Component}  if he is in the toolbox area and if he does not already hold one.
The robot can {\em Maintain} a device individually if it needs to be maintained and if in the same location.
Upon success, the device status turns to {\em good}.
    
    \item \uline{Rewards $R$:}
A reward of $+100$ is given when all devices have been repaired or maintained.
A reward (penalty) of $-2$ is given for each action except for the human's {\em Wait}, and a penalty of $-20$ is given in case of invalid action. A penalty of $-1$ is given if the human waits before having repaired all the broken devices.
If all devices are in "good" status, no penalty (0) is given to the human wait action.
The penalty associated to the wait action encourages the robot to postpone maintenance actions if this helps the human finish repair actions early. 
\end{itemize}

This large Dec-POMDP has 2\,304 states, 49 joint actions (7 actions per agent), and 5\,400 joint observations (30 for the human and 180 for the robot). Note: This state space grows quadratically with the number of cells.

To represent uncertainty about the human objectives, we replaced the reward function described previously by two variants:
\begin{enumerate*}
\item the human prefers to repair the broken device on the left first, receiving a $+10$ reward in that case; and
\item symmetrically, the human prefers to repair the broken device on the right first.
\end{enumerate*}
These objectives generate different human policies using \Cshref{ext-alg:MpomdpSampling}.
Initially, the robot only has a prior over the human's rewards and associated policy (each with probability $0.5$).
Due to the required coordination for repairing devices, these human policies have to account for the robot's ability to help the human when needed.

\subsection{Experiments with Synthetic Humans}

We used the method described in \Cshref{sec:Generate_human_policies} to compute the stochastic human FSCs associated to each human objective, and the method described in \Cshref{sec:robust_best_response} to compute the robust robot policy with the POMDP solver SARSOP \cite{sarsop}.
To save resources, an {\em action-threshold} (here always set to $0.1$) is used to prune low-probability human actions when computing the stochastic human FSCs.
We conducted experiments with
3 values for the softmax parameter $T$ ($0$, $0.3$ and $0.5$) and
several values for the maximum stochastic FSC size $N_{\max}$. 

\subsubsection{Qualitative Results}

{\em We first observed that the extracted human stochastic FSC can effectively encode possible trajectories of the human to solve the task if $N_{\max}$ is large enough for the current $T$.}
For instance, when $T = 0.3$ and $N_{\max} = 100$, the extracted human stochastic FSCs reach depths\footnote{The depth of an FSC is the maximum distance between the (here unique) initial node and any node.} of 22 and 18 (for each objective), which are sufficient for the human to finish the task if the robot collaborates.
However, when $N_{\max} = 50$, the depth covered by the FSC of the {\em prefer-left} objective is only 13 (15 for the 2nd objective), while a depth of 15 is required to finish the task.
We also observed that the higher the temperature $T$, the larger the $N_{\max}$ value needed for the human stochastic FSC to regularly finish the task.
When $T$ increases, more actions are considered at each node, generating more branches and preventing sufficient depth.
Note also that, even if we set $T$ to a very low value (near 0), the softmax distribution may contain several optimal actions. This allows  encoding several optimal human policies in a single FSC.

{\em When $N_{\max}$ is large enough for the current $T$, we observed that the computed robust robot policy 
can successfuly solve the task, accounting for the uncertainties over the human objectives and behaviors.}
It helps the human repair all the devices and performs the required maintenance operation.
For example, when $T = 0.3$, $N_{\max} = 100$, the robot following the generated robust policy first goes to cell $(0, 0)$ and waits for the human to pick a component at cell $(2,2)$.
Afterwards, if the robot observes that the human is approaching, it keeps waiting for him and then helps him repair the left device when the human reaches cell $(0,0)$. 
But if it observes the human moving to the top-right corner, then the robot decides to go right and help the human repair the right device first.
The robot then moves to cell $(1,0)$ to perform a maintenance operation while the human is going to pick a second component.
Finally, the robot helps the human repair the other broken device to finish the task.

Even if this seems like a simple pattern 
for the robot, 
it still requires the robot to reason on many possible human trajectories.
For example, if there are $2$ optimal actions in each state, even for 5 time steps, there are $2^5$ possible optimal trajectories (and this number can increase dramatically when considering sub-optimal actions).
Moreover, in this complex collaboration scenario, agents make
decisions only based on partial observations, and, as illustrated
above, the robot infers the human objective despite his uncertain
behavior.

\subsubsection{Robustness Analysis}

We also wanted to quantitatively analyze the robustness of our approach by comparing the method described in \Cshref{sec:Generate_human_policies} with a method computing a robot policy only based on deterministic human policies.
To that end,
we extracted 50 pairs of deterministic human policies, one per objective, using $T=0.5$ and $N_{\max}=600$  (\cf end of \Cshref{sec:Generate_human_policies}), but obtaining less than $200$ nodes in all cases; then,
for each pair, we computed a best-reponse robot policy to the equiprobable union of these two human policies (\cf beginning of \Cshref{sec:robust_best_response}). The resulting set of $3\times 50$ policies is thus ($\Pi_H =  \{ \langle \pi^1_{H,l}, \pi^1_{H,r}, \pi^1_{H,l+r} \rangle, \dots , \langle \pi^{50}_{H,l}, \pi^{50}_{H,r}, \pi^{50}_{H,l+r} \rangle\}$).
For each human union policy  $\pi^i_{H,l+r}$, we also compute a robot best response $\pi^i_{R,BR}$ by solving a POMDP obtained as in \Cshref{sec:robust_best_response}.

In our experiments, we compare the robust robot policies $\pi^*_{R}$ obtained for different values of $T$ and $N_{\max}$ against
\begin{enumerate*}
\item the 50 best responses $\pi^i_{R,BR}$ described above (averaging over all of them), and
\item the robot policy obtained by solving (with Inf-JESP \cite{YouThoColBuf-ictai2021}) the corresponding Dec-POMDP.\footnote{This solution assumes a strong a priori coordination since the human and robot agree on their individual policies.}
\end{enumerate*}
We measure the average value and success rate of each such robot policy against all $3\times 50$ synthetic human policies.

The obtained results, along with standard deviations, are presented in \Cref{tab:res_table}.
First, as expected, the worst results are obtained with the Dec-POMDP and Best-Response solutions ($\pi_{R,Dec-POMDP}$ and $\pi_{R,BR}$), with near-zero success rates.
Our approach improves a little bit in terms of success rate when using $T=0$, \ie, when assuming that the human only chooses among a small subset of actions at each time step,
which leads to fairly small human FSCs (24 and 23 nodes for the two independent objectives).
The average reward remains very low, though.
Using $T=0.3$ or $0.5$ leads to much better results.
In the {\em prefer-right} scenario, human FSCs with 200 nodes are even sufficient to get very good solutions (90\% success rate).
Moving from $T=0.3$ to $T=0.5$, this phenomenon is all the more important that more erratic behaviors require larger FSCs to better account for most likely trajectories.
In the {\em prefer-left} scenario, the success rates and values drop significantly.
This is due to the longer trajectories needed in this case, which in turn require large human FSCs to encode good enough policies.
Then, in the ``union'' scenario where the human's preference is randomly sampled, the results are a simple combination of the results in the {\em prefer-left} and {\em prefer-right} scenarios.

Note that, in these experiments, the synthetic humans used for evaluation are rather erratic ($T=0.5$), which explains that the best success rates are obtained with robot policies derived for that same temperature, while robots derived for $T=0.3$ are not robust enough.

\Cref{tab:time_results} presents the recorded computational time used in each step of our work.
The 1st step consists in converting each Dec-POMDP task model to a MPOMDP (one by human objective);
the 2nd step in generating stochastic human policies (FSCs) as presented in \Cshref{sec:Generate_human_policies} (the most time-consuming process);
the 3rd step in building the robot POMDP using the task models (Dec-POMDPs) and the human FSCs;
and, the final step in solving the robot POMDP using SARSOP to get a robust robot policy.
Here, most of the time is spent calling POMCP in the FSC extraction.

\begin{table*}\caption{Average value (and success rates in parentheses) of various robot policies vs. various human behaviors.
For each robust robot policy $\pi^*_R(T,N_{\max})$, we also indicate the sizes of both generated human FSCs as $N=(N_{left},N_{right})$.
}
  \label{tab:res_table}

\sisetup{
      detect-weight=true,detect-inline-weight=math, round-mode = places,
      round-precision = 1,
      table-format=-2.1
    }
  
    \setlength{\tabcolsep}{0pt}
  
\newcommand{\valCel}[2]{#1 & $\pm$ & #2 }
\newcommand{\pcCel}[1]{( & #1 & \% && ) ~}

    \centerline{
      \begin{adjustbox}{max width=\textwidth}
\begin{tabular}{clc
          ScS[table-format=2.2] c
          cS[table-format=2.0, round-precision=0]cS[table-format=1.0, round-precision=0]c c
          ScS[table-format=2.2] c
          cS[table-format=2.0, round-precision=0]cS[table-format=1.0, round-precision=0]c c
          ScS[table-format=2.2] c
          cS[table-format=2.0, round-precision=0]cS[table-format=1.0, round-precision=0]c c
          ScS[table-format=2.2] c
          cS[table-format=2.0, round-precision=0]cS[table-format=1.0, round-precision=0]c c
          }
          \toprule
&&
          & \multicolumn{29}{c}{$3\times 50$ synthetic human FSCs sampled using $T=0.5$}  & & \multicolumn{8}{c}{$10$ real humans} & \\
          \cmidrule(rl){4-32} \cmidrule(rl){34-42}
          ~ && ~
          & \multicolumn{9}{c}{$\pi^{Sample}_{H, l}$}
          & ~ & \multicolumn{9}{c}{$\pi^{Sample}_{H, r}$}
          & ~ & \multicolumn{9}{c}{$\pi^{Sample}_{H, l+r}$}
          & ~ & \multicolumn{9}{c}{$\pi^{Real}_H$} & ~ \\
          \cmidrule(lr){2-2} \cmidrule(lr){4-12} \cmidrule(lr){14-22} \cmidrule(lr){24-32} \cmidrule(lr){34-42}
          \rowcolor{gray!0}
          & $\pi_{R,Dec-POMDP}$ &
          & \valCel{-34.88}{6.69} && \pcCel{0.0}  && \valCel{-82.52}{14.35} & ~ & \pcCel{0.0} && \valCel{-58.70}{10.52}  & ~ & \pcCel{0.0} &&&&&&&&&&& \\
          \rowcolor{gray!8}
          & $\pi_{R,BR}$ &
          & \valCel{-172.104}{2.53}  &&  \pcCel{3.48}  && \valCel{-174.437}{2.54}  && \pcCel{3.36}  && \valCel{-173.25}{2.54}  && \pcCel{3.42} &&&&&&&&&&&  \\
           \rowcolor{gray!16}
          & $\pi^*_R(T=0.0,N_{max} =100, N = (\,24,\,23))$ &
          & \valCel{-109.99}{12.54}  && \pcCel{10.0}  && \valCel{-179.448}{15.42}   && \pcCel{18.00}  && \valCel{-144.72}{13.99}  && \pcCel{14.0}   && \valCel{-55.27}{6.53} & ~ & \pcCel{16.25} & \\
          \rowcolor{gray!0}
          & $\pi^*_R(T=0.3, N_{max} = 200, N = (200,200))$ &
          & \valCel{-38.06}{9.80}  && \pcCel{52.00}  && \valCel{\B 10.00}{2.00} && \pcCel{90.0} && \valCel{-14.03}{5.91} && \pcCel{71.0} &&&&&&&&&&& \\
           \rowcolor{gray!8}
          & $\pi^*_R(T=$ \lineOf{$0.3$}$, N_{max} = 400, N = (400,400))$ &
          & \valCel{-11.85}{7.15}  && \pcCel{68.00}   && \valCel{\B 10.00}{2.00} && \pcCel{90.0}  && \valCel{-0.921}{4.575}  && \pcCel{79.0} &&&&&&&&&&&  \\
           \rowcolor{gray!16}
          & $\pi^*_R(T=$ \lineOf{$0.3$}$, N_{max} = 600, N = (600,443))$ &
          & \valCel{-11.85}{7.15}  && \pcCel{68.00}   && \valCel{\B 10.00}{2.00}  && \pcCel{90.0} && \valCel{-0.925}{4.58} && \pcCel{79.0} && \valCel{\B -20.79}{7.35} &&  \pcCel{\B 81.25} & \\
          \rowcolor{gray!0}
          & $\pi^*_R(T=0.5, N_{max} = 200, N = (200,200))$ &
          & \valCel{-36.61}{5.58} && \pcCel{16.0}   && \valCel{\B 10.00}{2.00} && \pcCel{90.0}   && \valCel{-13.27}{3.78}  && \pcCel{53.0} &&&&&&&&&&& \\
            \rowcolor{gray!8}
          & $\pi^*_R(T=$ \lineOf{$0.5$}$, N_{max} = 400, N = (400,400))$ &
          & \valCel{2.655}{2.519} && \pcCel{76.00}   && \valCel{\B 10.02}{2.00}  && \pcCel{90.0}     && \valCel{6.34}{2.26} && \pcCel{83.00} &&&&&&&&&&& \\
            \rowcolor{gray!16}
          & $\pi^*_R(T=$ \lineOf{$0.5$}$, N_{max} = 600, N = (600,600))$ &
          & \valCel{\B 9.536}{2.179} && \pcCel{84.0}    && \valCel{\B 10.02}{2.00} && \pcCel{90.0}   && \valCel{\B 9.77}{2.08}  && \pcCel{\B 87.13} && \valCel{-33.66}{11.91} && \pcCel{75.00} & \\
          \bottomrule
        \end{tabular}
      \end{adjustbox}
      }
  \end{table*}

\begin{table}
  \caption{CPU Time (in seconds) for different experiments}
  \label{tab:time_results}

  \sisetup{
    round-mode = places,
    round-precision = 0,
    table-format=2,
    table-alignment = center
}
  
  \centering
{
    \begin{tabular}{lS@{ }S@{ }SS@{ }S@{ }S}
    \toprule

            & \multicolumn{3}{c}{$\pi^*_R(T, N_{\max})$} \\
    Step    & { ($0.0, 100$)} & { ($0.3, 600$)} & {($0.5, 600$)} \\
    \midrule
    Dec-POMDP$\to$MPOMDP            & 13    &     13   &  13         \\
    Get Human Stoc. FSCs            & 83    &   2032  &   6144     \\
    Build Robot POMDP               & 3     &   178     &  203   \\
    Solve Robot POMDP               & 2     &   57      &  105       \\
      \midrule
    Total time                      & 101   &   2280    &   6465      \\
    \bottomrule
    \end{tabular}
    }
\end{table}

\subsection{Real human Experiments}
  
  To further evaluate our approach, we implemented this collaboration task as a computer game where the human is controlled through the keyboard.
To make a friendly user interface, the human player can directly observe the state of the environment, as shown in \Cshref{Collaboration_Task}, but the robot still faces partial observability as defined in the Dec-POMDP problem, and behaves accordingly.
Then, we selected three robot policies, $\pi^*_R(0.0,100)$, $\pi^*_R(0.3,600)$ and $\pi^*_R(0.5,600)$, from the previous experiment,
and we invited 10 real human subjects to collaborate with those robot policies.
Each subject was informed that the goal of this collaboration task was to turn each device status to ``good'' within at most 30 time steps, but did not know or observe the instant reward.
He or she then played 8 consecutive rounds of this collaboration game per robot policy (for a total of 24 rounds), the order of the robot policies being randomly selected.

\subsubsection{Qualitative results}
We first observed that the robot policy $\pi^*_R(0.0, 100)$ was unable to help the human subjects since it only accounts for few high-quality human actions.
For example, a common mistake of the human subjects was to forget to pick a component necessary to repair a broken device.
To fix that mistake, the human subject had to go back to the toolbox, thus generating an unexpected human behaviour for robot policy $\pi^*_R(0.0, 100)$, and leading to a failure due to the robot policy getting stuck in a self-loop.
On the contrary, while no specific information regarding human behaviors were provided,  robot policies $\pi^*_R(0.3, 600)$ and $\pi^*_R(0.5, 600)$ managed to recover and help the human to solve the task because they consider more possible human actions in each situation.
In most cases, those policies adapted to different human behaviors and were able to tolerate human mistakes such as the one mentioned above.
  
Moreover, after experiments, we asked our subjects how they felt about the 3 tested robot policies.
$60\%$ of our subjects felt that there was no adaptation at all when playing with $\pi^*_R(0.0, 100)$, and the other $40\%$ felt that they needed to adapt to $\pi^*_R(0.0, 100)$ to finish the task.
On the other hand, $50\%$ of the subjects reported mutual adaptations between them and the two robot policies $\pi^*_R(0.3, 600)$ and $\pi^*_R(0.5, 600)$;
$30\%$ reported that $\pi^*_R(0.3, 600)$ and $\pi^*_R(0.5, 600)$ were adapting to their behaviors;
and
$20\%$ felt they needed to adapt to the robot.
Last but not least, we asked our subjects to choose one robot policy which makes they feel comfortable during the collaboration task, $60\%$ of our subjects chose $\pi^*_R(0.3, 600)$ and $40\%$ on  $\pi^*_R(0.5, 600)$, but no one choose the robot policy $\pi^*_R(0.0, 100)$.

\subsubsection{Quantitative results}
The average cumulative rewards and success rates of human subjects are shown in the right part of \Cref{tab:res_table}.
For the same three robot policies, the cumulative rewards are lower than with synthetic human experiments ($-55.3$, $-20.8$ and $-33.7$ respectively).
There are multiple reasons causing this drop of values: 
first, the human subjects needed some rounds to get familiar with the game, therefore, in early rounds, human usually make more mistakes and received penalties; 
moreover, since human subjects were not informed of the exact reward function, they were not aware of penalties (negative rewards) obtained when waiting or when going outside of the grid-world.
  
  On the other hand, the observed success rates of human subjects with robot policies $\pi^*_R(0.3, 600)$ and $\pi^*_R(0.5, 600)$ are $81\%$ and $75\%$ respectively.
This shows that, even if the cumulative rewards are low, or if the human performs mistakes, $\pi^*_R(0.3, 600)$ and $\pi^*_R(0.5, 600)$ could still help the human to accomplish the task most of the time.
The higher success rate (and cumulative reward) with $\pi^*_R(0.3, 600)$ than with $\pi^*_R(0.5, 600)$ (contrary to the results with synthetic humans obtained with $T=0.5$) suggests that the robot's mental model of humans is better when it builds human FSCs using $T=0.3$ than $T=0.5$.

\section{CONCLUSION}

In this paper, we address the problem of uncertainty over human's objectives in human-robot collaboration.
The contribution is twofold:
\begin{enumerate*}
\item First, we discuss the chicken-and-egg problem in the second-order mental model, and provide a method to overcome this obstacle and automatically generate a human FSC model which aggregates various possible behaviors for each human objective.
Parameters can be tuned to adjust the diversity of the generated human behaviors.
\item Second, we propose a robust robot planning algorithm that relies on a POMDP with uncertainties on the human's actual objective and behavior.
We formally detail how to build this robot POMDP based on the task models (Dec-POMDPs) and a distribution over human stochastic policies (FSCs), one per possible objective.
\end{enumerate*}
Note that the robust planning algorithm (2) does not depend on how the human policies are derived (1).
Also, the human policies only serve as the robot's mental model of the human (and to conduct some experiments);
actual humans may behave differently.

Through experiments, we demonstrate that our approach is robust to uncertain human behaviors with different objectives.
While our scenario considered the same task but with different human preferences, different tasks could be included in the same way.
We believe this work is important for collaboration settings where the robot and the human need to reason on each other's possible actions, and where considering myopic or deterministic human policies is not sufficient to generate robust robot policies.
Moreover,  our approach only requires a Dec-POMDP describing the human-robot collaboration task,  no prior human behavior being needed.
This makes our method generic to tackle different human-robot collaboration problems, the main issue being its scalability when facing large problems.
Future work will focus on scaling up the approach, and allowing to replace the exact Dec-POMDP model by a simulator, \eg, by relying only on simulation-based solvers.
We also plan to conduct further experiments, and, in particular, implement our approach on a real scenario
where a drone has to help a human to repair devices.

\renewcommand*{\bibfont}{\footnotesize}
\printbibliography 

\ifdefined\extended

\onecolumn

\appendix

\subsection{Details about the Human FSC Extraction}
\label{app|FscExtraction}

\Cref{alg:MpomdpSampling} details the process for extracting an FSC to model a human behavior from the optimal $Q$-value of M-POMDP $M$ (\cf \Cshref{sec:Generate_human_policies}).

First, a unique start node is created (\cshref{alg_part:StartNode}) with $\sigma^T_H(\cdot|b_0)$, $\sigma^T_R(\cdot|b_0)$, $b_0$ and the initial weight $w=1$.
Then $n_0$ is added to both the new FSC ($N$) and an open list ($G$) (\cshref{alg_part:AddInitNodeToN}). Getting inspiration from LAO* \cite{HANSEN200135}, we would like to expand at each iteration the node which has greatest contribution to $b_0$'s value.
Thus, while $G$ is not empty, 
we select the node $n \in G$ that has the highest estimated value $V^*(n.b)$, weighted by the probability to reach that node (estimated through $n.w$) 
(\cshref{alg_part:SortOpenList}),
then process that node with possible observation-action pairs under current belief $b$ (\cshref{alg_part:PopNode}--\ref{alg_part:FindAllHumanPossibleFscObs}), impossible observation-action pairs inducing a self-loop (\cshref{alg_part: SelfLoop}).

\Cref{alg_part:BeliefUpdate,alg_part:ComputeNewWeight} compute the updated belief $b'$
and its weight $w'$  (see detailed formulas below).
Before creating a new node $n'$ with those new computed components, we also need to check: 
\begin{enumerate}
\item if the new belief $b'$ does not already exist inside the FSC ($N$) considering a \textit{Norm-1} distance threshold $\epsilon$;
\item if the FSC does not already contain the maximum node size $N_{\max}$.
\end{enumerate}
If both conditions are satisfied (\cshref{alg_part:CheckValidNewNode}),  we create a new node $n'$ and add it to $N$.
Otherwise, we find and process the node $n'$ in $N$ which has the closest belief to the new computed belief $b'$.

\begin{algorithm}
  \caption{Extract a human stochastic FSC} \label{alg:MpomdpSampling}
  \DontPrintSemicolon
  
  \SetInd{.5em}{1em} 

  {

    {\bf [Input:]} $T$: softmax temperature $\mid$ \linebreak $N_{\max}$: max \# of nodes $\mid$ $\epsilon$: max belief gap \;
    
$n_0 \gets node( \angles{ \sigma_H^T( \cdot | b_0),  \sigma_{R}^T( \cdot | b_0), b_0, w=1 })$ \label{alg_part:StartNode}\; $ N \gets \{n_0\}$  \label{alg_part:AddInitNodeToN}  \, ; \,
    $ G.pushback(n_0)$  \label{alg_part:AddInitNodeToG}   \;
\While{$|G| > 0$}{
      $G.sort()$  \label{alg_part:SortOpenList} \tcp{\scriptsize sort nodes according to $w\cdot V^*(b)$}
      $n \equiv \angles{ \sigma_H^T, \sigma_{R}^T, b, w } \gets G.popfront() \label{alg_part:PopNode} $\;
\ForAll{ $\angles{ o_H, a_H } \in \Omega_H \times A_H$}	{
        \eIf{$Pr(o_H, a_H | b, \sigma_H^T)>0$ \label{alg_part:FindAllHumanPossibleFscObs} }
        {
          $b' \gets updateBelief(b,a_H,o_H,\sigma_{R}^T(\cdot|b) )$ \label{alg_part:BeliefUpdate} \;
$w' \gets w \cdot Pr(o_H, a_H | b, \sigma_H^T(\cdot|b), \sigma_{R}^T(\cdot|b) )$ \label{alg_part:ComputeNewWeight} \;
\eIf{$( b' \notin N(\epsilon) ) \wedge ( N.size() < N_{\max} ) $ \label{alg_part:CheckValidNewNode} }{
            $n' \gets node( \angles{ \sigma_H^T(\cdot|b'), \sigma_{R}^T(\cdot|b'), b', w' })$ \;   
            $N \gets N \cup \{n'\}$ \, ; \,
            $G.pushback(n')$ \;
            $\eta(n, \angles{ a, o }, n') \gets 1$\;
            
          }{
            $n' \gets N.findClosest(b')$\;
            $n'.w \gets   n'.w  + w' $\;
            $\eta(n, \angles{ a, o }, n') \gets 1$\;
            
          }
          
        }
        {
          $\eta(n, \angles{ a, o }, n) \gets 1$ \label{alg_part: SelfLoop} \;
        }
        
      }
    }
    \Return{$\angles{ N, \eta, \psi }$}
    
  }
\end{algorithm}

The belief update in \cref{alg_part:BeliefUpdate} is computed as follows:
\begin{align}
  b'(s')
  & = Pr(s' | b, a_H, o_H, \sigma_{R}^T(\cdot|b) )
   = \frac{ Pr(s', o_H | b, a_H, \sigma_{R}^T(\cdot|b) ) }
  { \sum_{a_H} Pr(s', o_H | b, a_H, \sigma_{R}^T(\cdot|b) )  }
  \\
  & \propto \sum_{s,a_R} Pr(s', o_H | s, \angles{a_H, a_R}) \cdot \sigma_{R}^T(a_R|b) \cdot b(s)
  \\
  & = \sum_{s,a_R,o_R} Pr(s', \angles{o_H, o_R} | \angles{a_H, a_R}) \cdot Pr(s' | s, \angles{a_H, a_R}) \cdot \sigma_{R}^T(a_R|b) \cdot b(s)
  \\
  & = \sum_{a_R} \left(\sum_{o_R} O( \angles{a_H, a_R}, s', \angles{o_H, o_R})\right) \cdot \left( \sum_s T(s, \angles{a_H, a_R}, s') \cdot b(s) \right) \cdot \sigma_{R}^T(a_R|b)
  .
\end{align}
As expected, it relies solely on the DecPOMDP's transition and observation functions, the previous belief $b$, and the robot's action sampling rule for that belief (and conditioned on the current human action).

Similarly, the probability used for updating weights in \cref{alg_part:ComputeNewWeight} is derived with
\begin{align}
  & Pr(o_H, a_H | b, \sigma_H^T(cdot|b), \sigma_{R}^T(\cdot|b) )
   = \sum_{a_R} Pr(o_H, a_H | b, \sigma_H^T(\cdot|b), \sigma_{R}^T(\cdot|b), a_R )
  \\
  & = \sum_{a_R,o_R,s,s'} Pr(s, s', \angles{ o_H, o_R }, a_H| b, \sigma_H^T(\cdot|b), \sigma_{R}^T(\cdot|b), a_R )
  \\
  & = \sum_{a_R,o_R,s,s'} b(s) \cdot Pr( s', \angles{ o_H, o_R } | s, \angles{ a_H, a_R }) \cdot \sigma_H^T(a_H|b) \cdot\sigma_{R}^T(a_R|b)
  \\
  & = \sum_{a_R,o_R,s,s'} b(s) \cdot Pr( \angles{ o_H, o_R } | s, \angles{ a_H, a_R }, s') \cdot Pr( s' | s, \angles{ a_H, a_R }) \cdot \sigma_H^T(a_H|b) \cdot\sigma_{R}^T(a_R|b)
  \\
  & = \sum_{a_R,o_R,s,s'} b(s) \cdot O( \angles{ a_H, a_R }, s', \angles{ o_H, o_R }) \cdot T( s, \angles{ a_H, a_R }, s') \cdot \sigma_H^T(a_H|b) \cdot\sigma_{R}^T(a_R|b)
  .
\end{align}

\subsection{Sampling a Deterministic FSC}
\label{app:samplingDeterministicFSC}

\Cref{alg:MpomdpSamplingDeterministic} samples a single deterministic human policy in the form of an FSC.
As shown in red, it differs from \Cref{alg:MpomdpSampling} only in that, in any node, a single human action is used, rather than a probability distribution over human actions.

\begin{algorithm}
  \caption{Sample a human deterministic FSC} \label{alg:MpomdpSamplingDeterministic}
  \DontPrintSemicolon
  
  \SetInd{.5em}{1em} 

  {

    {\bf [Input:]} $T$: softmax temperature $\mid$ \linebreak $N_{\max}$: max \# of nodes $\mid$ $\epsilon$: max belief gap \;
    
$n_0 \gets node( \angles{ \textcolor{red}{ a_H \sim \sigma_H^T( \cdot | b_0)},  \sigma_{R}^T( \cdot | b_0), b_0, w=1 })$ \label{alg2_part:StartNode}\; $ N \gets \{n_0\}$  \label{alg2_part:AddInitNodeToN}  \, ; \,
    $ G.pushback(n_0)$  \label{alg2_part:AddInitNodeToG}   \;
\While{$|G| > 0$}{
      $G.sort()$  \label{alg2_part:SortOpenList} \tcp{\scriptsize sort nodes according to $w\cdot V^*(b)$}
      $n \equiv \angles{ \textcolor{red}{ a_H }, \sigma_{R}^T, b, w } \gets G.popfront() \label{alg2_part:PopNode} $\;
\ForAll{ $o_H \in \Omega_H$}	{
        \eIf{$Pr(o_H, a_H | b, \sigma_H^T)>0$ \label{alg2_part:FindAllHumanPossibleFscObs} }
        {
          $b' \gets updateBelief(b,a_H,o_H,\sigma_{R}^T(\cdot|b) )$ \label{alg2_part:BeliefUpdate} \;
$w' \gets w \cdot Pr(o_H, a_H | b, \sigma_H^T(\cdot|b), \sigma_{R}^T(\cdot|b) )$ \label{alg2_part:ComputeNewWeight} \;
\eIf{$( b' \notin N(\epsilon) ) \wedge ( N.size() < N_{\max} ) $ \label{alg2_part:CheckValidNewNode} }{
            $n' \gets node( \angles{ \textcolor{red}{a_H \sim \sigma_H^T(\cdot|b')}, \sigma_{R}^T(\cdot|b'), b', w' })$ \;   
            $N \gets N \cup \{n'\}$ \, ; \,
            $G.pushback(n')$ \;
            $\eta(n, \angles{ a, o }, n') \gets 1$\;
            
          }{
            $n' \gets N.findClosest(b')$\;
            $n'.w \gets   n'.w  + w' $\;
            $\eta(n, \angles{ a, o }, n') \gets 1$\;
            
          }
          
        }
        {
          $\eta(n, \angles{ a, o }, n) \gets 1$ \label{alg2_part: SelfLoop} \;
        }
        
      }
    }
    \Return{$\angles{ N, \eta, \psi }$}
    
  }
\end{algorithm}

\subsection{Complementary Empirical Results}

When experimenting with human subjects, we considered 3 robot policies: $\pi^*_R(0.0, 100)$,  $\pi^*_R(0.3, 600)$ and $\pi^*_R(0.5, 600)$.
Each human subject needed to play 8 consecutive rounds of this collaboration game per robot policy,
and the order of robot policies were randomly selected in order to remove side effects due to the need of some rounds to get familiar with the task.
\Cref{fig:real_human_experiment_results} presents the results of the different sessions and the answers the human subject gave to our questions after each series of 8 rounds.

\Cref{fig:real_human_experiment_results} (a) presents each subject's number of successes for the collaboration task with the different robot policies.
Since each subject encountered each robot policy 8 times, a value of 8 means that this human subject managed to solve the task each round while a value of 0 means this human subject never managed to solve the task while collaborating with this robot policy.
When human subjects tried to collaborate with $\pi^*_R(0.0, 100)$,
$30\%$ of the subjects (3 out of 10) never succeeded to solve the collaboration task,
another $30\%$ succeeded only once, and
$40\%$ succeeded 2 or 3 times.
This shows that this robot policy does not adapt to the spontaneous human behavior, and that the human subjects have difficulties to find a behavior that would trigger appropriate robot reactions.
However, when human subjects collaborated with robot policies $\pi^*_R(0.3, 600)$ and $\pi^*_R(0.5, 600)$, they all succeeded in more than 4 rounds, 
and some of them even successful accomplished all 8 rounds with one of those two robot policies.
These similar results suggests that increasing temperature $T$ from $0.3$ to $0.5$ does not improve the robot's robustness to actual human behaviors.

\Cref{fig:real_human_experiment_results} (b) presents the detailed subjects' evaluation of both players' adaptation during the experiments.
After each series of 8 rounds, we asked the human subjects to assess the interaction with the robot.
Each human subject had to pick one choice representing his or her impression among the following four choices: ``\textit{no adaptation}'', ``\textit{human adapts to robot}'', ``\textit{robot adapts to human}'', and ``\textit{mutual adaptation}''.
When playing with $\pi^{*}_{R}(0.0, 100)$, $60\%$ of our subjects reported that there was no adaptation at all, and the other $40\%$ felt that they needed to adapt to the current policy to finish the task.
On the other hand, we observed the subjects gave the same distribution of choice for the robot policies $\pi^{*}_{R}(0.3, 600)$ and $\pi^{*}_{R}(0.5, 600)$:
$50\%$ of the subjects reported mutual adaptation;
$30\%$ reported that $\pi^{*}_{R}(0.3, 600)$ and $\pi^{*}_{R}(0.5, 600)$ were adapting to their behaviors;
and $20\%$ felt they needed to adapt to the robot.
This confirms that it makes little difference for the humans to play with these last two robots, while their results where significantly different in experiments with (erratic) synthetic humans (designed with $T=0.5$).

\Cref{fig:real_human_experiment_results} (c) indicates which robot was preferred by the human subjects after all $3\times 8$ rounds of this collaboration game (thus having played with all three robot policies).
As expected by the previous results, no one picked robot policy $\pi^{*}_{R}(0.0, 100)$ since it was hard to collaborate with the robot and the interactions often led to a failure of the task, as shown in \Cref{fig:real_human_experiment_results} (a). 
Then, $60\%$ of our subjects picked $\pi^{*}_{R}(0.3, 600)$ and $40\%$ $\pi^{*}_{R}(0.5, 600)$.
The fact that $\pi^{*}_{R}(0.3, 600)$  was the most selected policy could imply that computing the robot policy based on overly erratic human models might be counterproductive (since the robot might be seen as being too indecisive), as also highlighted by the success rate presented in \Cref{tab:res_table}. 
However, more subjects and more investigations are needed to verify that assumption.

\begin{figure}
  \centering
  \subfigure[]{\includegraphics[width=0.32\textwidth]{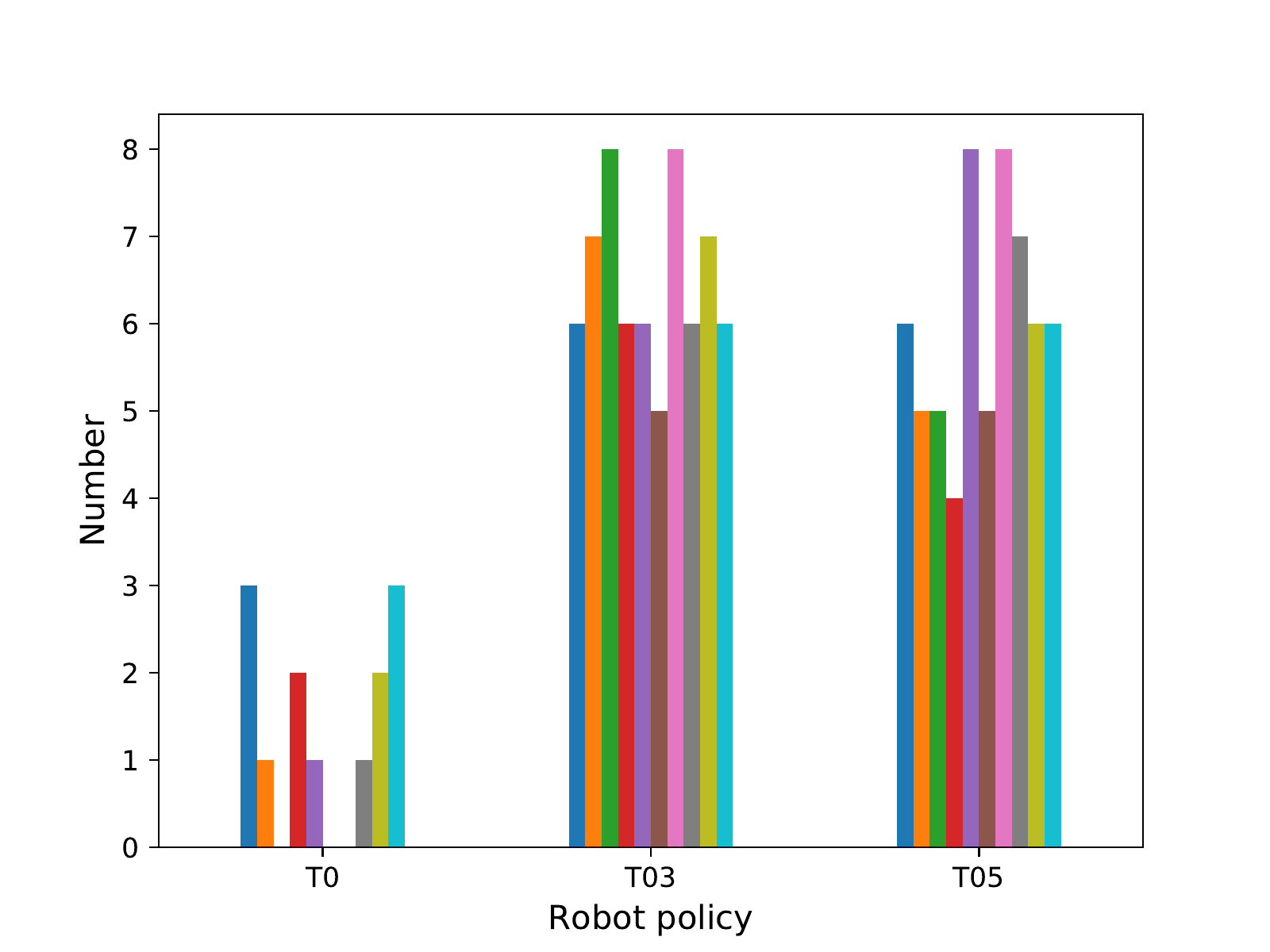}} 
  \subfigure[]{\includegraphics[width=0.32\textwidth]{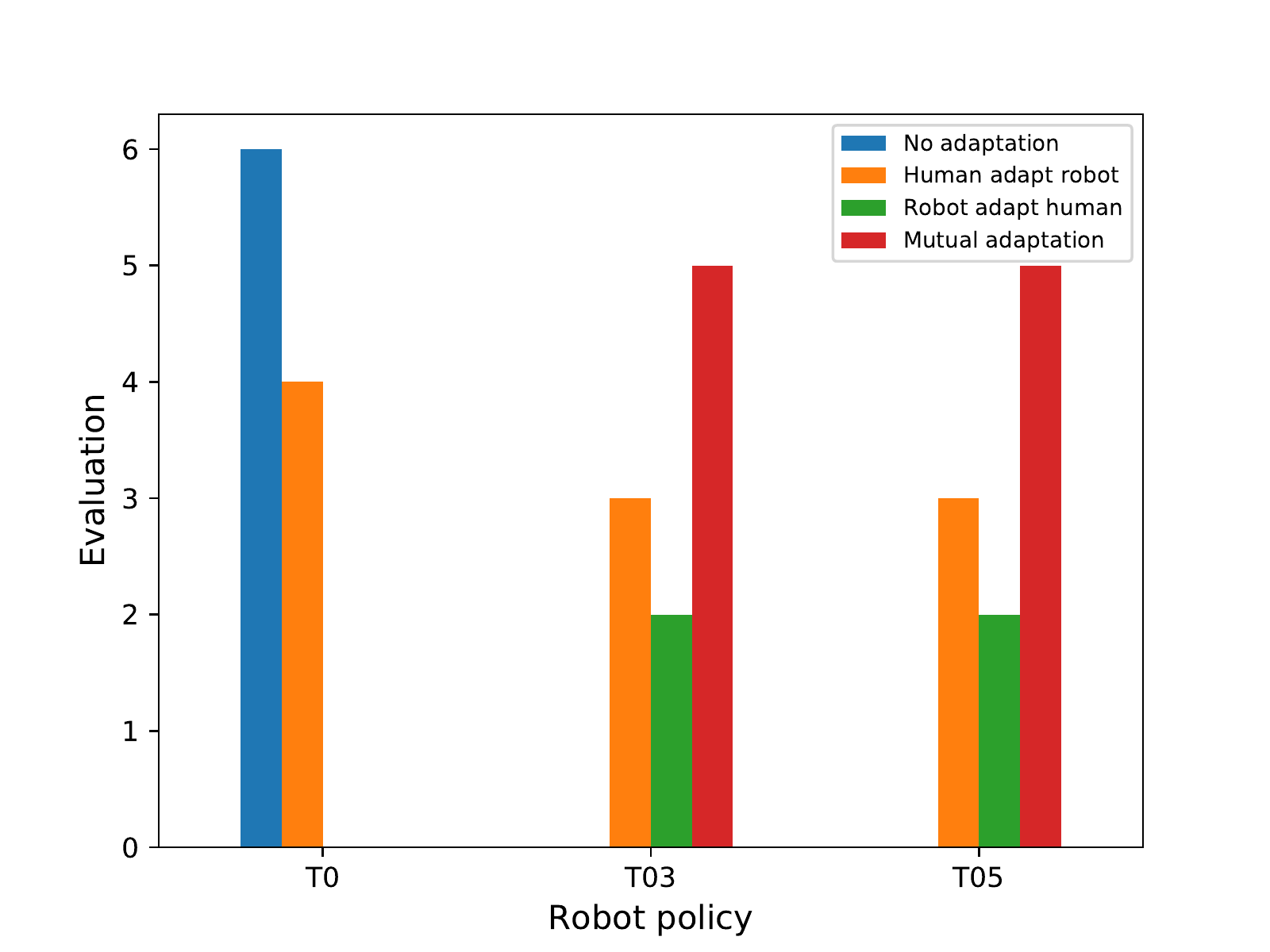}} 
  \subfigure[]{\includegraphics[width=0.32\textwidth]{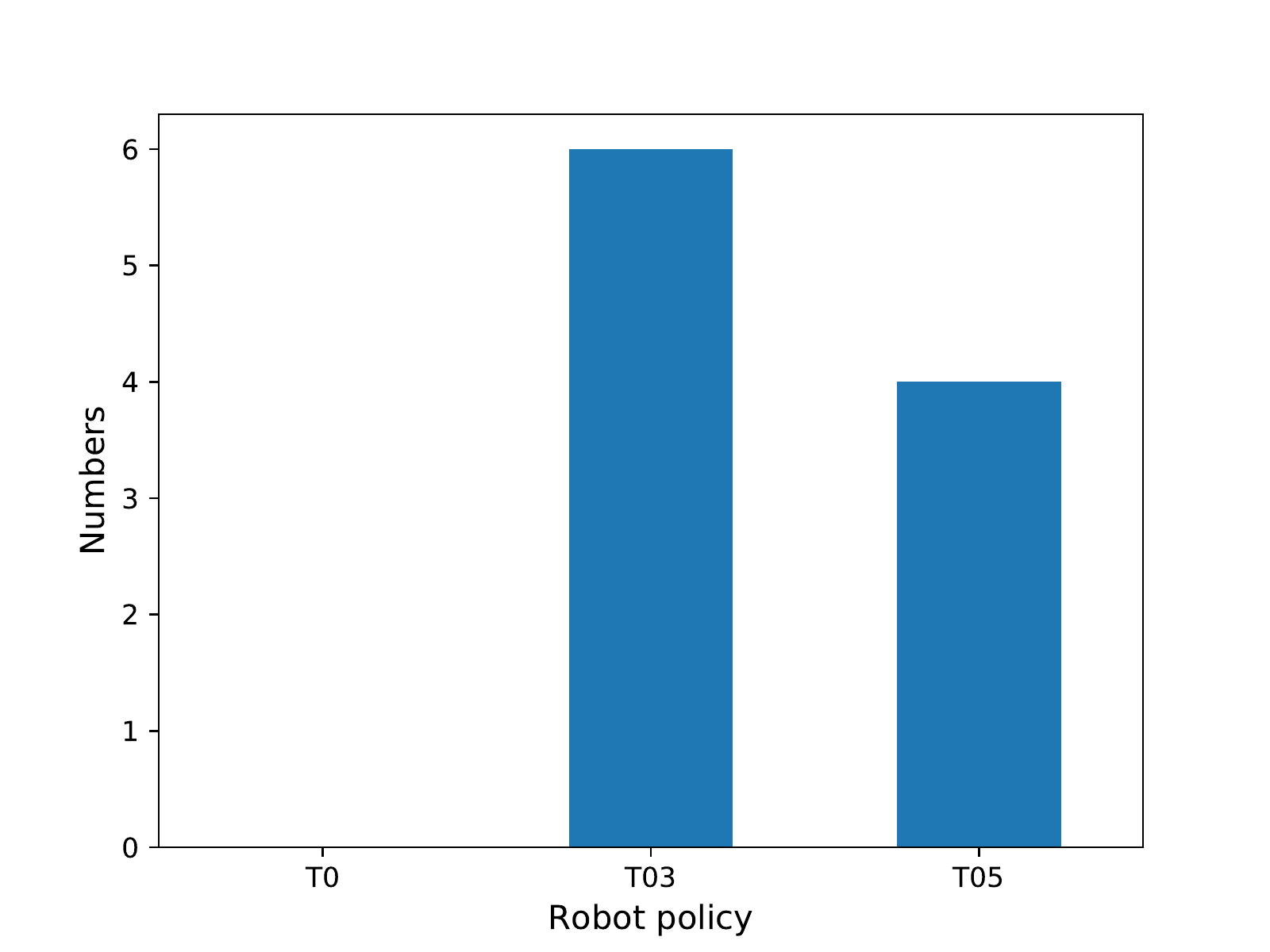}}
  \caption{Figure~(a) presents each subject's number of successes when he or she collaborates with different robot policies.
Figure~(b) provides, for each robot policy, all subjects' evaluations regarding both players' adaptation during the collaboration task.
Figure~(c) shows the subjects' choices when asked to pick the robot policy they preferred at the end of the experiments.}
  \label{fig:real_human_experiment_results}
\end{figure}

 \fi

\end{document}